\pdfoutput=1

\documentclass[11pt]{article}

\usepackage{ACL2023}

\usepackage{times}
\usepackage{latexsym}

\usepackage[T1]{fontenc}

\usepackage{graphicx}
\usepackage{subfigure}
\usepackage{color}
\usepackage{tabularray}

\usepackage[utf8]{inputenc}

\usepackage{nicefrac}

\usepackage{microtype}

\usepackage{inconsolata}

\usepackage{xcolor}

\usepackage{amsmath}
\usepackage{array}
\usepackage{multirow}
\usepackage{color}
\usepackage{geometry}
\usepackage{setspace}
\usepackage{caption}
\usepackage{booktabs}

\usepackage{amsmath}  
\usepackage{amssymb}  

\newcommand{\comment}[1]{}

\title{Characterizing Stereotypical Bias from Privacy-preserving Pre-Training}

\author{Stefan Arnold \and Rene Gröbner \and Annika Schreiner \\ 
Friedrich-Alexander-Universität Erlangen-Nürnberg \\ Lange Gasse 20, 90403 Nürnberg, Germany \\ 
\texttt{(stefan.st.arnold, rene.edgar.gröbner, annika.schreiner)@fau.de}}

\begin{document}

\maketitle

\begin{abstract}

Differential Privacy (DP) can be applied to raw text by exploiting the spatial arrangement of words in an embedding space. We investigate the implications of such text privatization on Language Models (LMs) and their tendency towards stereotypical associations. Since previous studies documented that linguistic proficiency correlates with stereotypical bias, one could assume that techniques for text privatization, which are known to degrade language modeling capabilities, would cancel out undesirable biases. By testing \texttt{BERT} models trained on texts containing biased statements primed with varying degrees of privacy, our study reveals that while stereotypical bias generally diminishes when privacy is tightened, text privatization does not uniformly equate to diminishing bias across all social domains. This highlights the need for careful diagnosis of bias in LMs that undergo text privatization. 

\end{abstract}

\section{Introduction}

\textit{Language Models} (LMs) \citep{devlin2019bert, radford2019language} are trained on large corpora of text that may contain confidential information. Since such information can be recovered from word embeddings \citep{song2020information, thomas2020investigating} and language models \citep{carlini2019secret, nasr2023scalable}, privacy emerged as an active concern for building trust and complying with stringent regulations on privacy protection.

To protect against unintended disclosure of information, \textit{Differential Privacy} (DP) \citep{dwork2006calibrating} has been integrated into machine learning \citep{abadi2016deep} and language models \citep{mccann2017learned,shi2022selective,du2023dp}. DP formalizes privacy through a notion of indistinguishability so that the model outputs are not affected by the addition or removal of an entry in the training corpus. This is accomplished by injecting additive noise on gradients during model training. 

Due to scaling issues associated with DP on LMs during perturbation of per-sample gradient updates \citep{abadi2016deep}, there is a trend towards perturbing the raw text \citep{fernandes2019generalised, feyisetan2020privacy, yue2021differential, chen2023customized}. 

By exploiting the geometric proximity of words in word embeddings \citep{mikolov2013efficient}, \citet{feyisetan2020privacy} proposed a probabilistic mechanism grounded in metric DP \citep{chatzikokolakis2013broadening} to perturb all words in a text while ensuring plausible deniability \citep{bindschaedler2017plausible} of the text regarding its provenance and content.

However, several studies documented that mechanisms for embedding words in a high-dimensional space harbor \citep{bolukbasi2016man, caliskan2017semantics, garg2018word, manzini2019black} and transfer \citep{papakyriakopoulos2020bias} unwanted stereotypes and prejudices present in a text corpus.

\paragraph{Contribution.} Building on the rich body of research exploring privacy-fairness trade-offs \citep{bagdasaryan2019differential, farrand2020neither, petren2022impact}, this study addresses the implications of text privatization on biased associations in LMs. Specifically, we pre-train \texttt{BERT} \citep{devlin2019bert} models with masked language modeling and next sentence prediction on webscraped text modified under varying levels of privacy. We then score the stereotypical bias following the context association test of \citet{nadeem2021stereoset} and stereotype pairs benchmark of \citet{nangia2020crows}. Our findings reveal a nuanced landscape where stereotypical bias generally diminishes as privacy guarantees are tightened. This is in line with prior research indicating that LMs with impaired language modeling capabilities tend to exhibit less stereotypical associations \citep{nadeem2021stereoset}. However, this diminution is not uniform across all social categories as biases associated with certain attributes show varying trends of stability, amplification, and attenuation. We thus advocate for careful bias measurement when deploying privacy-preserving LMs.



\section{Background}
\label{sec:background}

To ensure a consistent understanding of privacy and fairness in machine learning, we provide the foundations of differential privacy and a brief definition of stereotypical bias along with related work.

\subsection{Differential Privacy}

Differential Privacy (DP) \citep{dwork2006calibrating} originated in the field of statistical databases and was adapted to machine learning \citep{abadi2016deep}. DP formalizes privacy through the indistinguishability of model outputs with respect to the presence or absence of a record in the dataset. The notion of indistinguishability is achieved through noise and can be controlled by the privacy budget $\varepsilon \in (0,\infty]$, with privacy guarantees diminishing as $\varepsilon \rightarrow \infty$.


Despite evidence of preventing information disclosure, the perturbations caused by noise can have detrimental \citep{jayaraman2019evaluating} and disparate \citep{bagdasaryan2019differential, farrand2020neither, petren2022impact} effects on the behavior of machine learning models. By assessing the accuracy of differentially private machine learning models for (underrepresented) subgroups, \citet{bagdasaryan2019differential} find a disparate impact regarding gender and ethnicity in both vision and text.

To prevent the risk of authorship disclosure, text rewriting is an appealing strategy that applies noise at word level or sentence level by leveraging word embeddings \citep{mikolov2013efficient} or sequence-to-sequence models \citep{vaswani2017attention}. Each approach comes with distinct mechanisms and implications for balancing utility and privacy.

\paragraph{Embedding-based Text Rewriting.} 

\citet{feyisetan2020privacy} pioneered a mechanism for text rewriting termed \texttt{Madlib}. \texttt{Madlib} exploits the distance of words in embedding spaces \citep{mikolov2013efficient} to substitute all words in a text with another word within a radius controlled by the privacy budget $\varepsilon$. Since this substitution mechanism scales the notion of indistinguishability by a distance, it satisfies the axioms of metric DP \citep{chatzikokolakis2013broadening}.

Building on a word embedding, the substitution involves three steps at word level: (1) retrieving the continuous representations of words from the embedding space, (2) adding noise to the representations calibrated using a multivariate distribution, and (3) mapping the noisy representation back onto the discrete space of vocabulary by employing a nearest neighbor approximation. While the probabilistic nature of these substitutions assures plausible deniability \citep{bindschaedler2017plausible}, substitutions based on the distance between words alleviate the curse of dimensionality typical of randomized response \citep{warner1965randomized}.

However, privatizing text through perturbations at word level imposes notable limitations. Since the privacy guarantee in this approach depend on the geometry of the embedding space, it necessitates meticulous calibration of the noise magnitude \citep{xu2020differentially}. For dense regions of the embedding space, excessive noise may obscure suitable substitutions. For sparse regions of the embedding space, minimal noise may not provide sufficient protection against reconstruction. In addition the to noise calibration, perturbations at word level, albeit retaining the meaning of a text, encounter difficulties in maintaining the coherence of the text, such as grammar \citep{mattern2022limits}, ambiguity \citep{arnold2023driving}, and hierarchy \citep{feyisetan2019leveraging}.

\paragraph{Autoencoder-based Text Rewriting.}

Instead of privatization over word embeddings, an orthogonal approach utilizes sequence-to-sequence models built on recurrent \citep{bo2021er, krishna2021adept, weggenmann2022dp} and transformer \citep{igamberdiev2023dp} architectures. Common to these approaches is that noise is added to the encoder representations of text and the decoder learns to convert these noisy representations into text but without stylistic identifiers. 

By perturbing the text at sentence level, this approach presents unique challenges compared to perturbing texts at word level. For instance, \citet{igamberdiev2022dp} criticized that the utility is contingent upon the resemblance between the texts on which the sequence-to-sequence model was optimized and the texts that are subjected to privacy-preserving paraphrasing. This limitation in generalizability renders this form of text rewriting infeasible for the privatization of pretext at scale.


\subsection{Stereotypical Bias}

Bias in machine learning is viewed as prior information that informs algorithmic learning \citep{mitchell1980need}. When the prior information is predicated on stereotypes and prejudices, bias transcends this neutral definition and manifests in a disproportionate weight in favor of or against a social group.


The origins of these problematic biases are often rooted in the raw data used to develop machine learning models \citep{caliskan2017semantics}. Implicit or explicit stereotypes based on characteristics such as gender and race can cause the models to perpetuate and propagate these biases. This can significantly affect perception and decision making. The issue with stereotypical bias is particularly acute in the context of language models due to their extensive training on vast corpora that reflect biases present in human language. This bias magnifies the potential to influence its tone \citep{dhamala2021bold} and content \citep{abid2021persistent}, resulting in negative effects on individuals and society at large.

Using tests for association analogies, prior research demonstrated that embeddings harbor stereotypical biases related to gender \citep{bolukbasi2016man, kurita2019measuring, chaloner2019measuring} and race \citep{manzini2019black}. Specifically, \citet{caliskan2017semantics} showed that terms related to career are associated with male names rather than female names, whereas unpleasant terms are associated with ethnic minorities. \citet{garg2018word} elaborate on the temporal dimension of bias in word embeddings by observing changes in gender and ethnic stereotypes over a century. This diachronic analysis indicates that while certain stereotypes have diminished over time, others remain robustly encoded in language. By investigating bias diffusion, \citet{papakyriakopoulos2020bias} showed that biases contained in word embeddings can permeate natural language understanding, while \citet{abid2021persistent} report stereotypes in language generation such as violence for certain religious groups. 

Unlike these studies on bias in raw data, we examine the bias that stems from text privatization. 

\section{Methodology}
\label{sec:methodlogy}

To test our hypothesis on amplification of stereotypical bias through text privatization, we need to define (1) a language model, (2) the mechanism for text privatization, and (3) a bias measurement. 

\subsection{Language Model} 

Following \citet{qu2021natural}, we use a \texttt{BERT} model \citep{devlin2019bert} leveraging masked language modeling and next sentence prediction tasks for pre-training. The choice of \texttt{BERT} is motivated by its widespread adoption and proven effectiveness in capturing contextual relationships within text. 

For pre-training, we selected a webscraped replication of \texttt{WebText} \citep{radford2019language}, which compared to \texttt{WikiText} \citep{merity2016pointer}, covers a broader spectrum of topics, styles, and viewpoints. This diversity renders \texttt{WebText} particularly suited for examining the transfer of stereotypical biases from the pre-text corpus. For fine-tuning, we reproduced the experiments of \citet{bagdasaryan2019differential} but found no stereotypical bias other than a disparate impact due to sampling bias. 


To assess the alterations in stereotypical bias by text privatization, we trained a \texttt{BERT} model devoid of any privacy interventions, serving as a control to score amplification and attenuation, and three additional copies of the \texttt{BERT} model under varying degrees of privacy guarantees. Since all \texttt{BERT} models are identical in terms of architecture and optimization (differing solely in the degree of text privatization), this setup warrants a controlled comparison that isolates the effects of text privatization on the anchoring of stereotypical bias.



\subsection{Text Privatization} 

To privatize the \texttt{WebText} corpus, we operationalize the \texttt{Madlib} mechanism developed by \citet{feyisetan2019leveraging} for text privatization at word level. \texttt{Madlib} necessitates the utilization of continuous representations supplied by a word embedding. We integrate \texttt{Madlib} with \texttt{GloVe} \citep{pennington2014glove}. \texttt{GloVe} supplies a $400000$-words vocabulary, each mapped to a $300$-dimensional representation. The choice of \texttt{GloVe}  is motivated by the richness of its semantic space, making it an ideal candidate for privacy-preserving text privatization.

Since the privacy guarantee of \texttt{Madlib} is rooted in metric DP, we need to calibrate the noise parameter $\varepsilon$ according to the metric space of \texttt{GloVe}. This calibration involves an estimation of the plausible deniability \citep{bindschaedler2017plausible} through two proxy statistics \citep{feyisetan2020privacy}:

\begin{itemize}

    \item $N_w = \mathbb{P} \{ M(w) = w \}$ measures the number of \textit{identical} words that stem from perturbing a word given a privacy budget $\varepsilon$. We estimate $N_w$ by counting the occurrence of unaltered words after querying a random subset of $10000$ words for a total of $1000$ times.
    
    \item $S_w = |\mathbb{P} \{ M(w) = w^{'} \}|$ measures the number of \textit{unique} words that stem from perturbing a word given a privacy budget $\varepsilon$. We estimate $S_w$ by calculating the effective support of a word after querying the same random subset of $10000$ words for a total of $1000$ times.
    
\end{itemize}

We can relate the proxy statistics to the privacy budget. Adding more noise corresponds to a tighter privacy guarantee. This is indicated by a smaller value for $\varepsilon$ and results in a diverse set of perturbed words (low $N_w$ and high $S_w$). Adding less noise reflects a weaker privacy guarantee. This is characterized by a larger value for $\varepsilon$ and results in more frequent unperturbed words (high $N_w$ and low $S_w$).

\begin{figure}[!t]
    \subfigure[$N_w$ refers to the number of perturbed words that are \textit{identical} to a queried word.]
    {
        \includegraphics[width=0.48\textwidth]{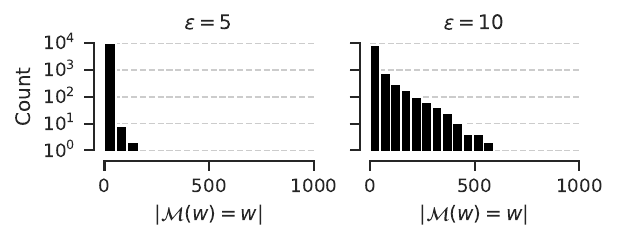}
        \label{fig:nw}
    }
    \hfill
    \subfigure[$S_w$ refers to the number of perturbed words that are \textit{unique} from a queried word.]
    {
        \includegraphics[width=0.48\textwidth]{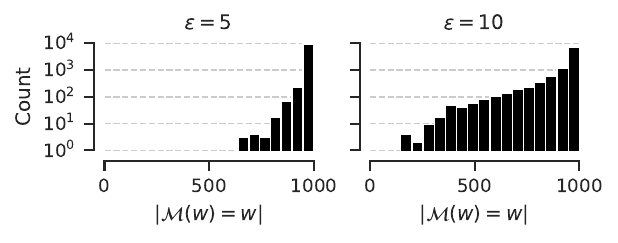}
        \label{fig:sw}
    }
    \caption{
        Plausible deniability statistics approximated for a randomly compiled vocabulary of $10000$ words,  each word privatized over a number of $1000$ queries.
    }
    \label{fig:privacy}
\end{figure}

\begin{table}
\footnotesize
\centering
\caption{Example sentence derived from \texttt{Webtext} and privatized for three independent runs of \texttt{Madlib} \citep{feyisetan2020privacy} using a privacy budget $\varepsilon$ of $10$.}
\label{tab:example}
\resizebox{\linewidth}{!}{%
\begin{tblr}{
  width = \linewidth,
  colspec = {Q[269]Q[644]},
  vlines,
  vline{1,3} = {-}{0.08em},
  hline{1,48} = {-}{0.08em},
  hline{2} = {-}{0.05em},
}
\textbf{Tokens} & \textbf{Substiutions}                        \\
Port-au-Prince  & \textit{rosita, xiangfan, tejgaon}           \\
,               & \textit{and, as, ,}                       \\
Haiti           & \textit{vanuatu, cuba, haiti}                \\
(               & \textit{(, 45, according}                    \\
CNN             & \textit{informed, journalist, speaker}       \\
)               & \textit{–, ), 2000}                          \\
–               & \textit{likely, –, two}                        \\
Earthquake      & \textit{quake, earthquake, stress}           \\
victims         & \textit{killings, murdered, deaths}          \\
,               & \textit{agrees, things, went}                \\
writhing        & \textit{desolation, stayers ,tiredness}      \\
in              & \textit{out, in, first}                      \\
pain            & \textit{frustration, fractures, pain}        \\
and             & \textit{have, over, with}                    \\
grasping        & \textit{interplay, spit, dangling}           \\
at              & \textit{at, the, as}                         \\
life            & \textit{proud, day, loves}                   \\
,               & \textit{and, took, 45}                       \\
watched         & \textit{watched, lined, raised}              \\
doctors         & \textit{medical, researchers, surgeons}      \\
and             & \textit{including, as, alongside}            \\
nurses          & \textit{pharmacists, nurses, physicians}     \\
walk            & \textit{walks, sideways, walked}             \\
away            & \textit{gone, away, when}                    \\
from            & \textit{from, around, off}                   \\
a               & \textit{an, than, one}                       \\
field           & \textit{games, yards, field}                 \\
hospital        & \textit{school, nursing, staff}              \\
Friday          & \textit{week, thursday, saturday}            \\
night           & \textit{night, hours, watch}                 \\
after           & \textit{after, afterwards, before}           \\
a               & \textit{a, first, one}                       \\
Belgian         & \textit{danish, macedonian, french}          \\
medical         & \textit{medical, hospital, psychiatric}      \\
team            & \textit{division, helm, cup}                 \\
evacuated       & \textit{evacuated, ferried, homeless}        \\
the             & \textit{the, 1984, on}                       \\
area            & \textit{town, area, park}                    \\
,               & \textit{accused, 6, :}                       \\
saying          & \textit{asking, iranians, saying}            \\
it              & \textit{since, as, is}                       \\
was             & \textit{that, only, subsequently}            \\
concerned       & \textit{suspicious, expect, insist}          \\
about           & \textit{nearly, just, about}                 \\
security        & \textit{beijing, actions, personnel}         \\
.               & \textit{still, then, .}            
\end{tblr}
}
\end{table}

Figure \ref{fig:privacy} presents the distribution of $N_w$ and $S_w$. Since $N_w$ ($S_w$) should be positively (negatively) skewed to assure a reasonable privacy guarantee, we adopt privacy budgets of $\varepsilon = \{5,10\}$, corresponding to a high and low level of privacy protection, respectively. Table \ref{tab:example} illustrates an example obtained by querying \texttt{Madlib} using a privacy budget $\varepsilon$ of $10$. Notice the fidelity while some variation asserts compliance with privacy requirements.

\subsection{Bias Measurement}

Characterizing bias embedded within models typically relies on carefully crafted datasets. Several datasets exist to measure bias in word embeddings \citep{caliskan2017semantics, may2019measuring} and language models trained with masked \citep{nangia2020crows, nadeem2021stereoset} and causal language modeling objective \citep{dhamala2021bold}. 


We adopt the \texttt{StereoSet} dataset designed by \citet{nadeem2021stereoset}. Given associative contexts, this dataset is intended to measure the tendency to default to stereotypical or anti-stereotypical associations. \texttt{StereoSet} provides meticulously crafted stimuli for bias measurement regarding gender, profession, race, and religion at two distinct levels:


\paragraph{Intrasentence.} The intrasentence task measures bias for sentence-level reasoning. It is formulated as a fill-mask task. Given a context sentence describing a social group, the task is to fill in a masked attribute corresponding to a stereotype, an anti-stereotype, and an unrelated option. The propensity for stereotypical associations is gauged by the likelihood of assigning each of these options.

\paragraph{Intersentence.} The intersentence task measures bias for discourse-level reasoning. It is formulated as a next-sentence task. Given a context sentence pertaining to a social group, followed by three sentences embodying a stereotype, an anti-stereotype, and an unrelated attribute, the assessment of stereotypical bias hinges on which of these sentences is instantiated as the most likely continuation.

To capture social biases at more differentiated levels, we complement our investigation with the \texttt{CrowS-Pairs} benchmark designed by \citet{nangia2020crows}. This benchmark consists of pairs of minimally distant sentences dealing with bias about gender identity, ethnic affiliation, age, nationality, religion, sexual orientation, socioeconomic status, physical appearance, and disability. The first sentence in each pair demonstrates a stereotype about a social group, while the second sentence in each pair violates it. This allows to score the bias in a language model by measuring how frequently it prefers a statement that portrays a social group stereotypically compared to an alternative portrayal of the same situation with a different social identity. 

Despite some criticism due to issues with model calibration \citep{desai2020calibration}, we determine the preferences using pseudo-likelihood scoring \citep{salazar2020masked}. We iterate over each sentence, masking a word at a time (except for the words that identify a social group), and accumulate the log-likelihoods of the masks in a sum for comparison.


\section{Experiments}

Prior to initiating our bias measurement, we conducted a preliminary sanity check by examining the pseudo-perplexity scores of \texttt{BERT} models trained under varying degrees of privacy. Pseudo-perplexity serves an indicator of a LM’s ability to accurately model the probability distribution of words within a text corpus, thereby reflecting the model’s proficiency to comprehend the linguistic structures encountered during its training.

We use a 10\% subset of \texttt{WikiText} for computing the pseudo-perplexities. Evaluated at privacy levels specified by the privacy parameter $\varepsilon$, the pseudo-perplexity scores were 93.51 with no privacy interventions, 502.67 with moderate privacy settings, and 2056.43 under conditions of high privacy. Consistent  with previous evidence that introducing noise at word-level compromises the linguistic proficiency of LMs \citep{mattern2022limits}, these results demonstrate a substantial  degradation as the level of privacy augmentation increases. 

The observed degradation raises an interesting question of whether private LMs harbor stereotypical biases despite diminished language modeling capabilities. This question forms the basis for our subsequent analysis of the undesirable biases in LMs stemming from text privatization.

\subsection{Stereotype Results from StereoSet}

\begin{table}
\footnotesize
\caption{Percentage preference of stereotypical associations derived from \texttt{StereoSet}, where scores above 0.5 indicate pro-stereotypical bias and scores below 0.5 indicate anti-stereotypical bias. Effect sizes compared to the baseline value according to Cohens $d$ in brackets.} 
\label{ss_stereoset}
\centering
\begin{tblr}{
  width = \linewidth,
  colspec = {Q[250]Q[150]Q[300]Q[300]},
  row{1} = {c},
  row{2} = {c},
  row{8} = {c},
  cell{2}{1} = {c=4}{0.937\linewidth},
  cell{8}{1} = {c=4}{0.937\linewidth},
  hline{1,14} = {-}{0.08em},
  hline{2-3,8-9} = {-}{0.05em},
  hline{7,13} = {-}{},
}
\textbf{Epsilon}       & $\infty$    & \textbf{10}                                   & \textbf{5}                             \\
\textbf{Intrasentence} &                &                                               &                                        \\
Gender                 & \textbf{.6196} & .5490 (\small{$\downarrow .14$})                 & .5020 (\small{$\downarrow .24$})          \\
Race                   & \textbf{.6060} & .5135 (\small{$\downarrow .19$})                 & .4709 (\small{$\downarrow .27$})          \\
Religion               & .5897          & \textbf{.6538} (\small{$\uparrow .13$})        & \textbf{.6538} (\small{$\uparrow .13$}) \\
Profession             & \textbf{.6062} & .5679 (\small{$\downarrow .08$})                 & .5259 (\small{$\downarrow .16$})          \\
Average                & \textbf{.6054} & .5711 (\small{$\downarrow .07$})                    & .5382 (\small{$\downarrow .14$})          \\
\textbf{Intersentence} &                &                                               &                                        \\
Gender                 & .5868          & \textbf{.5909} (\small{$\uparrow .01$})        & .5248 (\small{$\downarrow .12$})          \\
Race                   & .5318          & .5287 (\small{$\downarrow .01$})                 & \textbf{.5461} (\small{$\uparrow .03$}) \\
Religion               & \textbf{.5641} & .5513 (\small{$\downarrow .03$})                 & .5385 (\small{$\downarrow .05$})          \\
Profession             & \textbf{.6070} & .5272 (\small{$\downarrow .16$})                 & .4813 (\small{$\downarrow .25$})          \\
Average                & \textbf{.5724} & .5495 (\small{$\downarrow .05$})                 & .5227 (\small{$\downarrow .10$})          
\end{tblr}
\end{table}

To measure the bias resulting from text privatization at sentence and discourse level, we commence our analysis by detailing the stereotype scores derived from the \texttt{StereoSet} benchmark. The stereotype score is defined by the percentage of examples for which the LM assigns a higher probability to the pro-stereotypical word as opposed to the anti-stereotypical word. As such, scores closer to 0.5 are indicative of unbiased associations.

Table \ref{ss_stereoset} presents the averaged stereotype scores grouped by intrasentence and intersentence tasks and segmented by social categories \footnote{Since \texttt{Madlib} involves a probabilistic mechanisms, one could argue that the bias patterns of  the privacy budget $\varepsilon$ on social categories is caused by the randomness of text privatization. To test whether the observed patterns stem from randomness, we reproduced all experiments using three distinct seeds. The variance across different configurations suggests that these patterns are inherent to the privatization process and not merely artifacts of random perturbations.}. Several key trends inform our understanding of the impact of text privatization on stereotypical bias. We observe that results from the intrasentence task aligns with those from the intersentence task, showing that the stereotype scores decline as the privacy level intensifies. For the intrasentence tasks, the averaged stereotype scores decreased from 0.6054 to 0.5711 and 0.5382 as the privacy budget was tightened to 10 and 5, respectively. For the intersentence tasks, the stereotype scores decreased similarity from 0.5724 to 0.5495 and 0.5227, respectively. However, the fall in stereotype scores is overall more pronounced in the intrasentence task than in the intersentence task. This disparity implies that mask language modeling is affected more acutely than next sentence prediction, which requires a broader context to build stereotypical association. 

While text privatization generally reduces stereotypical biases, we find inconsistent pattern when breaking down the stereotype scores by social categories. This indicates that the impact of text privatization is not uniformly spread across social groups.

\subsection{Stereotype Results from CrowS-Pairs}

\begin{table}
\footnotesize
\caption{Percentage preference of stereotypes derived from \texttt{CrowS-Pairs}, where scores closer to 0.5 are indicative of unbiased associations. Effect sizes of text privatization compared to the baseline value in brackets.}
\label{ss_crowspairs}
\centering
\resizebox{\linewidth}{!}{%
\begin{tblr}{
  width = \linewidth,
  colspec = {Q[250]Q[150]Q[300]Q[300]},
  column{even} = {c},
  column{3} = {c},
  hline{1,11} = {-}{0.08em},
  hline{2} = {-}{0.05em},
}
\textbf{Epsilon} & \textbf{$\infty$} & \textbf{10}                            & \textbf{5}                             \\
Gender           & .5229           & \textbf{.5878}~(\small{$\uparrow .13$}) & .5267~(\small{$\uparrow .01$})          \\
Age              & .4943           & .4943~(\small{$\uparrow .00$})          & \textbf{.5402}~(\small{$\uparrow .09$}) \\
Race             & .5233           & .5446~(\small{$\uparrow .04$})          & \textbf{.5640}~(\small{$\uparrow .08$}) \\
Religion         & \textbf{.6000}  & .5905~(\small{$\downarrow .02$})          & .5905~(\small{$\downarrow .02$})          \\
Nationality      & .5283           & \textbf{.5535}~(\small{$\uparrow .05$}) & .5346~(\small{$\uparrow .01$})          \\
Occupation       & \textbf{.5465}  & .5407~(\small{$\downarrow .01$})          & .4535~(\small{$\downarrow .19$})          \\
Sexuality        & \textbf{.6786}  & .6190~(\small{$\downarrow .12$})          & .5119~(\small{$\downarrow .34$})          \\
Disability       & \textbf{.6167}  & .6000~(\small{$\downarrow .03$})          & .5500~(\small{$\downarrow .13$})          \\
Appearance       & .4762           & \textbf{.6190}~(\small{$\uparrow .29$}) & .4921~(\small{$\uparrow .03$})          
\end{tblr}
}
\end{table}

To explore the manifestation of stereotypical bias across a broader range of social categories, we broadened our analysis to include \texttt{CrowS-Pairs}. Table \ref{ss_crowspairs} confirms that there is no overarching trend regarding the degree of text privatization and the manifestation of stereotypical biases.

Following the general observation of decreasing stereotype scores as the privacy budget tightens, further scrutiny into social categories reveals a complex and heterogeneous response to text privatization. We discern social categories that are constant (e.g., religion), amplified (e.g., age, race), and attenuated (e.g., occupation, sexuality, disability). This suggests that some social categories are detached from the influences of textual perturbations while others seem less robust. Further complicating the interactions is that some social categories (e.g., gender, nationality, appearance) experience fluctuating responses. The categories show an increase in stereotype scores as privacy settings are intensified before stabilizing or reverting at the strictest levels of privacy. Except for sexual orientation ($\downarrow .34$) and physical appearance appearance ($\uparrow .29$), the effect sizes are negligible. This variability underscores the intricate dynamics between text privatization and LMs, suggesting that minor modifications in the privacy parameters can have significant and diverse impacts on stereotypical biases across different social constructs.

\section{Conclusion}

The interaction dynamics that govern the manifestation of bias in LMs are equivocal \citep{petren2022impact}. Prior research indicates that stereotypical bias is related to language proficiency in LMs \citep{nadeem2021stereoset}. Since text privatization is known to impair language modeling capabilities \citep{feyisetan2020privacy}, one would expect a general diminution of stereotypical bias. However, the word embeddings used for text privatization are documented to harbor \citep{bolukbasi2016man, caliskan2017semantics} and transfer \citep{papakyriakopoulos2020bias} stereotypical biases. This duality raises questions about whether text privatization leads to an amplification or an attenuation of stereotypical biases. By probing a LMs tendency to default to stereotypical or anti-stereotypical associations, we aimed to elucidate the relationship between text privatization and the amplification or attenuation of biases. We find that different social domains react differently to privacy settings and recommend to carefully assess stereotypical bias after training a LM on a privatized corpus of text.

\section{Limitations}

This study has several limitations that warrant consideration. Our experiments are based on \texttt{WebText}. While this corpus provides a broad range of topics and styles, it is possible that the derived insights, such as the general reduction in stereotypical bias and the unequal reduction across social groups, are influenced by spurious correlations \citep{schwartz2022limitations} inherent in the dataset. In addition to the flaws caused by the training corpus, our reliance on \texttt{GloVe} embeddings for text privatization introduces another potential source of inherent biases. Future research should address these limitations by incorporating a more diverse set of datasets and explore how alternative embeddings affect the persistence of stereotypical bias after privatization.

\bibliography{anthology,custom}
\bibliographystyle{acl_natbib}

\end{document}